\documentclass[conference]{IEEEtran}
\IEEEoverridecommandlockouts
\usepackage{cite}
\usepackage{amsmath,amssymb,amsfonts}
\usepackage{algorithmic}
\usepackage{graphicx}
\usepackage{textcomp}
\usepackage{xcolor}
\def\BibTeX{{\rm B\kern-.05em{\sc i\kern-.025em b}\kern-.08em
    T\kern-.1667em\lower.7ex\hbox{E}\kern-.125emX}}

\usepackage{booktabs}
\usepackage{lipsum}

\usepackage{subfig}
\newsavebox{\measurebox}
\usepackage[]{graphicx}
\usepackage{tikz}
\usepackage{siunitx}
\usepackage{booktabs}
\usepackage[shadow,textsize=scriptsize, textwidth=1.4cm]{todonotes}

\begin{document}

\title{Classification of Beer Bottles using Object Detection and Transfer Learning\\

}

\author{\IEEEauthorblockN{Philipp Hohlfeld\IEEEauthorrefmark{1},
Tobias Ostermeier\IEEEauthorrefmark{2},
Dominik Brandl\IEEEauthorrefmark{3} }
\IEEEauthorblockA{Faculty of Electrical Engineering and Information Technology,
Technical University of Applied Sciences Regensburg\\
Regensburg 93053, Germany\\
Email: \IEEEauthorrefmark{1}philipp.hohlfeld@st.oth-regensburg.de,
\IEEEauthorrefmark{2}tobias.ostermeier@st.oth-regensburg.de,\\
\IEEEauthorrefmark{3}dominik1.brandl@st.oth-regensburg.de}}

\maketitle

\begin{abstract}
Classification problems are common in Computer Vision. Despite this, there is no dedicated work for the classification of beer bottles. As part of the challenge of the master course Deep Learning, a dataset of \num{5207} beer bottle images and brand labels was created. An image contains exactly one beer bottle.
In this paper we present a deep learning model which classifies pictures of beer bottles in a two step approach. As the first step, a Faster-R-CNN detects image sections relevant for classification independently of the brand. In the second step, the relevant image sections are classified by a ResNet-18. The image section with the highest confidence is returned as class label. We propose a model, with which we surpass the classic one step transfer learning approach and reached an accuracy of \SI{99.86}{\percent} during the challenge on the final test dataset. We were able to achieve \SI{100}{\percent} accuracy after the challenge ended.

\end{abstract}


\section{Introduction}

Deep learning has become a buzzword nowadays due to its increasing popularity in recent years. The high availability of large datasets as well as powerful graphics processing units (GPUs) and high efficiency algorithms made it possible to achieve excellent results in many areas, such as image classification, object detection and natural language processing \cite{b6}.
Therefore in the elective master course Deep Learning at the Technical University of Applied Sciences Regensburg in the summer semester 2021, a Deep Learning Challenge (DLC) was organized for the first time.
The aim of this competition was to let the students compete in groups to solve a deep learning task in the best possible way.
The goal of the DLC 2021 was to classify beer bottles from different local breweries and to achieve the highest possible recognition rate.

The paper is structured as follows: Section \ref{Related Work} shows related work where similar problems have already been solved. Section \ref{Dataset} gives an overview of the dataset and the distribution of images. Section \ref{Proposed networks} presents an overview of the basic structure of the networks used. Section \ref{Experiments} describes our used datasets and models. In section \ref{Results} we present our results and in the last section \ref{Conclusion} we give a brief overview of our paper.
  
\section{Related Work} \label{Related Work}
Beer bottle logo classification can be seen as a subset of logo classification. For general image classification and especially for logo classification and detection there are various machine learning approaches.

Iandola et al. \cite{Iandola.2015} applies several different convolutional neural networks (CNNs) for logo classification and a pipeline made up of a fast region-based convolutional neural network (R-CNN) and different CNNs for logo detection. Iandola et al. achieves \SI{89.6}{\percent} accuracy with ``GoogLeNet-GP'' on FlickrLogos-32-test dataset.

Eggert et al. \cite{Eggert.2015} uses selective search to determine object proposals, propagates each proposal through a pretrained VGG16 net and classifies the obtained feature vectors with a series of Support Vector Machines (SVMs).

Oliveira et al. \cite{Oliveira.2016} applies a Fast-R-CNN to retrieve several object proposals per image. Each image is classified according to the top confidence region by a CNN.

Tran et al. \cite{Tran.2019} classifies close-up images of beer bottle caps with a VGG based custom model. The final model achieves a test accuracy of \SI{99.33}{\percent} on their custom beer bottle cap dataset.

We aim to build a model achieving the highest accuracy on the test dataset. Therefore we solve the problem of image classification in two steps. In the first step, image sections relevant for classification are detected independently of the logo. In the second step, the relevant image sections are classified. The image section with the highest confidence is returned as class label.

\section{Dataset} \label{Dataset}
The dataset provided for this purpose was made by the students themselves. At least \num{100} pictures were taken of each beer bottle. It is important that only one bottle is shown in a picture. Care was taken to always change the background, lighting and perspective to achieve a high variance in the images. Additionally, the beer types (e.g. hell, wheat beer or pilsner) of the same brewery differed in the pictures to make the DLC more challenging. 

The dataset contains \num{44} different classes. These images were divided into validation data, training data and two test datasets.
The distribution of the images is as follows: \num{2076} training images, \num{536} validation images, \num{1046} test  images and \num{1549} final test images.
The training and validation images were for training and finetuning the models and the test images were to check the achieved accuracy.
Additionally there was a final test dataset where the final result for the DLC of each model is determined.
The final test images were only available after the various models had been submitted.
This ensures that the trained network cannot be finetuned for the final test dataset.

\section{Proposed Methods} \label{Proposed networks}
Our basic idea to solve this challenge was using transfer learning. The main idea behind transfer learning is to extract knowledge from some related domains to help a machine learning algorithm to achieve greater performance in the domain of interest, in our case to classify beer bottles \cite{b1}. 

\subsection{Convolutional Neural Network}
A CNN is a type of deep learning model for processing data that has a grid pattern and designed to automatically and adaptively learn spatial hierarchies of features, from low- to high-level patterns. The great advantage of CNNs is that the features of the objects do not have to be determined in advance, but are learned by the network itself, without giving a definition of which feature exactly has to be searched for.
A CNN is typically composed of three types of layers: convolution, pooling, and fully connected layers \cite{b1000}.

The convolution layer is the core of CNN. Each of the convolution layers contains multiple filters, and each filter consists of multiple neurons. The convolution layer scans the image through the convolution kernel and fully utilizes the information of the adjacent regions in the image to extract the image features.\cite{b2}

\subsection{Deep Residual Network}
\begin{figure}[!h]
\centering
\includegraphics[keepaspectratio, width=0.5\textwidth] {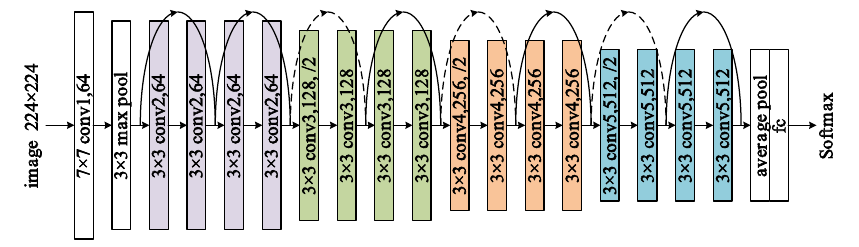} 
\caption{Network structure of ResNet-18 \cite{b2}}
\label{fig:Platzhalter2}
\end{figure}

The Deep Residual Network (ResNet) is one of the most commonly CNNs \cite{b2}. The pretrained ResNet-18 \cite{resnetPaper}, which has already proven itself in several classification and object detection tasks \cite{b2} \cite{b3} \cite{b4} was used in our experiments.
The network structure of ResNet-18 can be seen in figure  \ref{fig:Platzhalter2}.

The images from the dataset must first be normalized and scaled from $1024\times1024$ to the input image size $224\times224$ of ResNet-18.
ResNet-18 consists of \num{16} convolution layers, and several downsampling and fully connected layers. 
The kernel size of the first convolution layer is $7\times7$, all others are $3\times3$.
After average pooling the feature map of the last convolution layer, an eigenvector is obtained by full connection. Then the classification probability is obtained by normalization with softmax.
Based on the highest probability, the last fully connected layer assigns the input to one of the selected classes \cite{b2}.

\subsection{Faster-R-CNN}
A Faster-R-CNN is composed of two different modules. The first module is a deep fully convolutional network that suggests regions. The second module consists of a detector, which uses the proposed regions \cite{b2000}. 
Instead of developing a Faster-R-CNN framework from scratch, we used the free available Detectron2 \cite{detecron2}, which is Facebook AI Research's next-generation software system that implements state-of-the-art object detection algorithms \cite{b5}. 
After recognizing a certain class within a proposed region, it is labeled and a bounding box is drawn around the area.
Since this detector is computationally intensive, the execution of the Faster-R-CNN was completed on the Pro version of Google Colaboratory.

\subsection{Support Vector Machine}
Along with classical deep learning, we tried to use SVMs as a simple machine learning method to classify the images. A SVM maps the input into a higher dimensional feature space, in which the data should be linearly separable. The separation hyperplane ensures a high level of generalization. For multi-class classification one SVM is trained per class. \cite{Cortes.1995} \cite{Chapelle.1999}

The goal of the set of SVMs is to get a simple and fast to train baseline for our classification pipeline. From each cropped image of dataset 2 (see subsection \ref{dataset 2}), the feature vector is extracted through ResNet-101 \cite{ResnetV2.He.2016} pretrained on the ILSVRC-2012-CLS dataset \cite{ilsvrc-2012-cls}. The extracted features with shape $ 1\times2048 $ are used as input for a series of SVMs with a radial basis function kernel.

\section{Experiments} \label{Experiments}
In this section we first describe the datasets used for training and testing our networks. Then we present not only the best, but all of the network architectures we designed during the DLC.

\subsection{Used datasets}
\subsubsection{Dataset 1}
The first dataset contains all the unedited images provided, which are described in section \ref{Dataset}.
\subsubsection{Dataset 2} \label{dataset 2}
\begin{figure}[htb!]
	\centering
	\begin{minipage}[c]{.5\linewidth}
		\centering
		\subfloat[Original image]
		{\tikz[remember picture]\node[inner sep=0pt,outer sep=0pt] (a){
				\includegraphics[height=4cm,keepaspectratio]{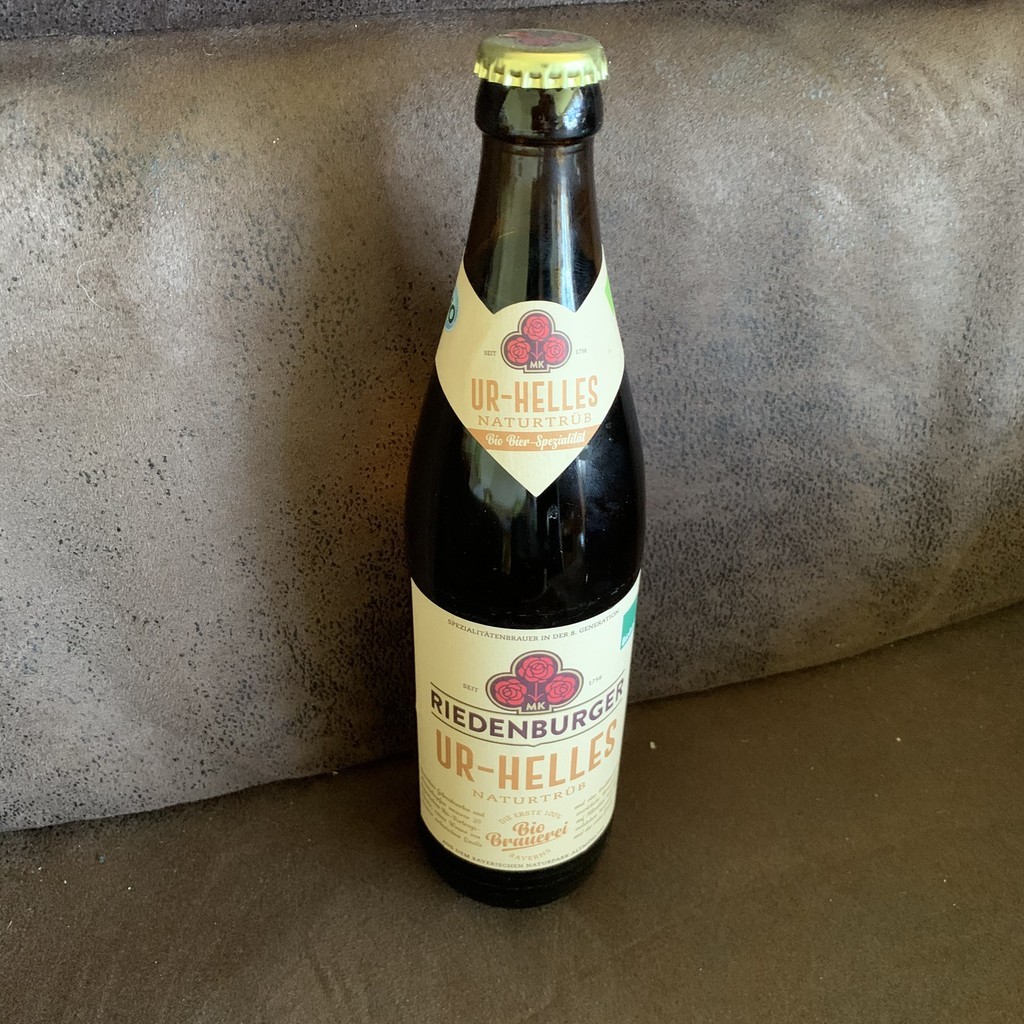}};
		}%
	\end{minipage}%
	\hfill%
	\begin{minipage}[c]{.5\linewidth}
		\centering
		\subfloat[Cropped 1]
		{\tikz[remember picture]\node[inner sep=0pt,outer sep=0pt] (b){
				\includegraphics[width=\linewidth,height=2.5cm,keepaspectratio]{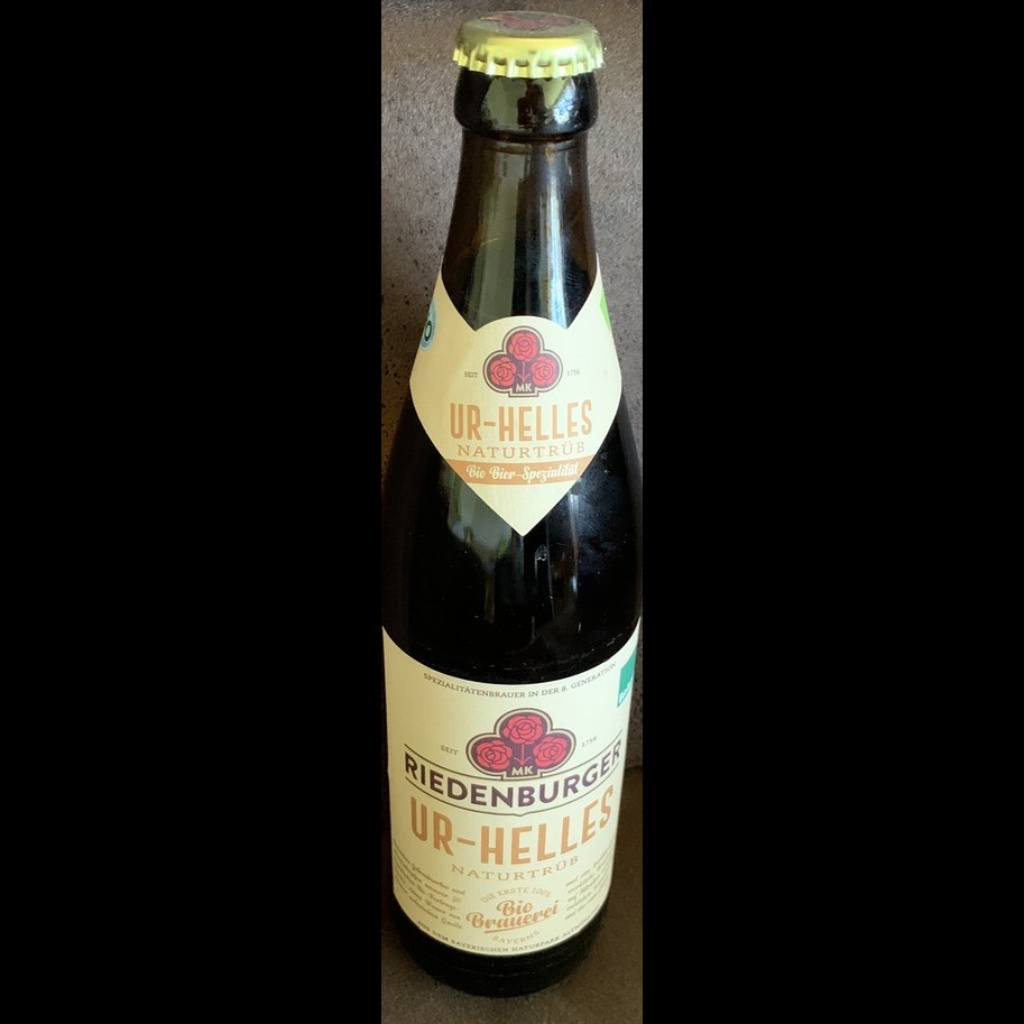}};
		}\\
		\centering
		\subfloat[Cropped 2]
		{\tikz[remember picture]\node[inner sep=0pt,outer sep=0pt] (c){
				\includegraphics[width=\linewidth,height=2.5cm,keepaspectratio]{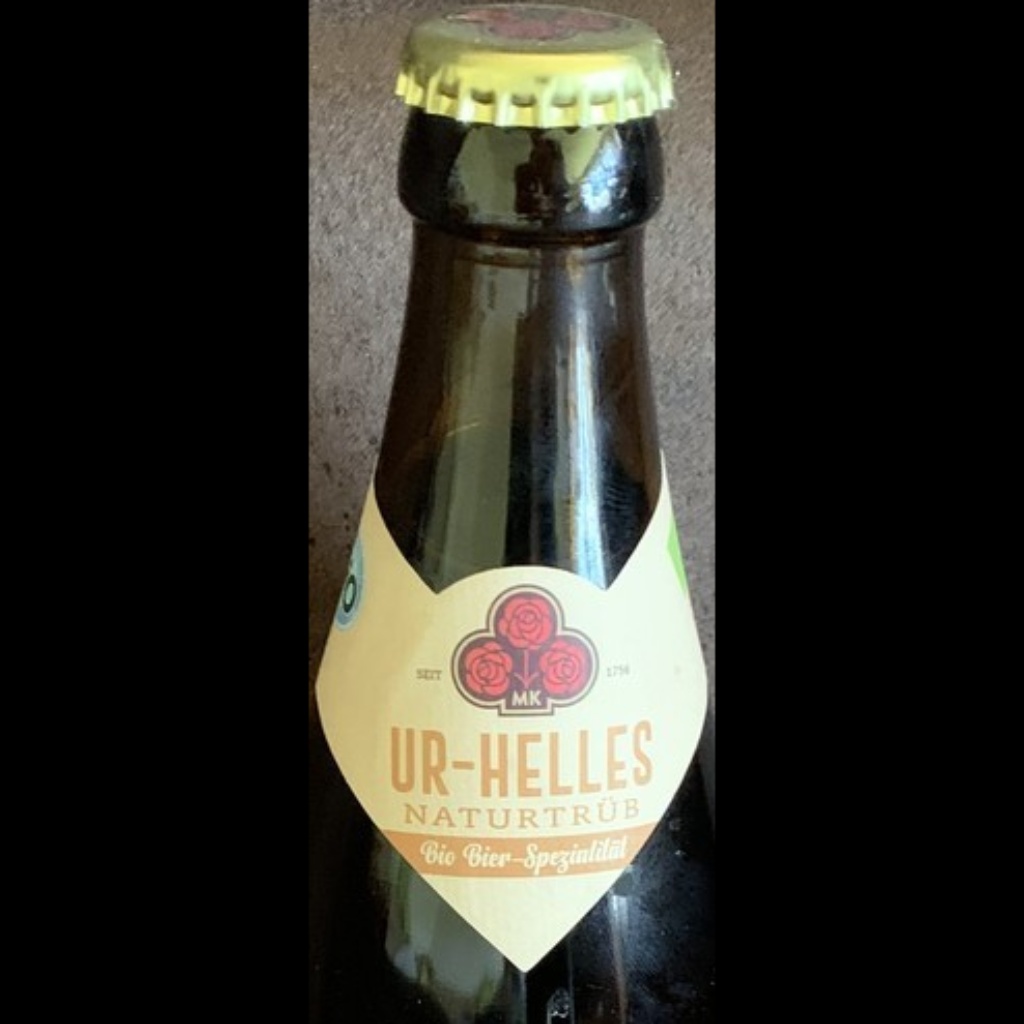}};
		}%
	\end{minipage}%
	\tikz[remember picture,overlay]\draw[line width=2pt,-stealth] ([xshift=2mm]a.east) -- ([xshift=-2mm]b.west);
	\tikz[remember picture,overlay]\draw[line width=2pt,-stealth] ([xshift=2mm]a.east) -- ([xshift=-2mm]c.west);
	
	\caption{Example of generating images for dataset 2 \label{fig:dataset2}}
  
\end{figure}

A second dataset was generated from the first one using the Detectron2 framework and a Faster-R-CNN pretrained on the COCO dataset. This Faster-R-CNN included a bottle class and was used with a low threshold value to find all bounding boxes of bottle like objects in an image. The bounding boxes were cropped from the original image, square padded to prevent distortion and resized to the original image size. Using a higher threshold resulted in missing cropped bottles when the beer bottle was not in the main focus of the image. False positives were sorted out manually. The final dataset had a total of \num{3925} images. \num{300} more than the original dataset because we kept all cropped images with at least part of the beer bottle label in it (see figure \ref{fig:dataset2}).

\subsubsection{BeerBB-1K dataset}
For training of the Faster-R-CNN the public available BeerBB-1K dataset was used \cite{beer1kgithub}. It contains \num{1000} images of beer bottles form different brands of various size and color and in different positions. Additionally the coordinates of the bottle bounding boxes are provided. For our network we dismissed the information of the brand and used it only to finetune the Faster-R-CNN on beer bottle recognition.



\subsection{Trained networks}
\subsubsection{Model 1}
\begin{figure}[h!]
	\centering
	\includegraphics[width=0.5\textwidth]{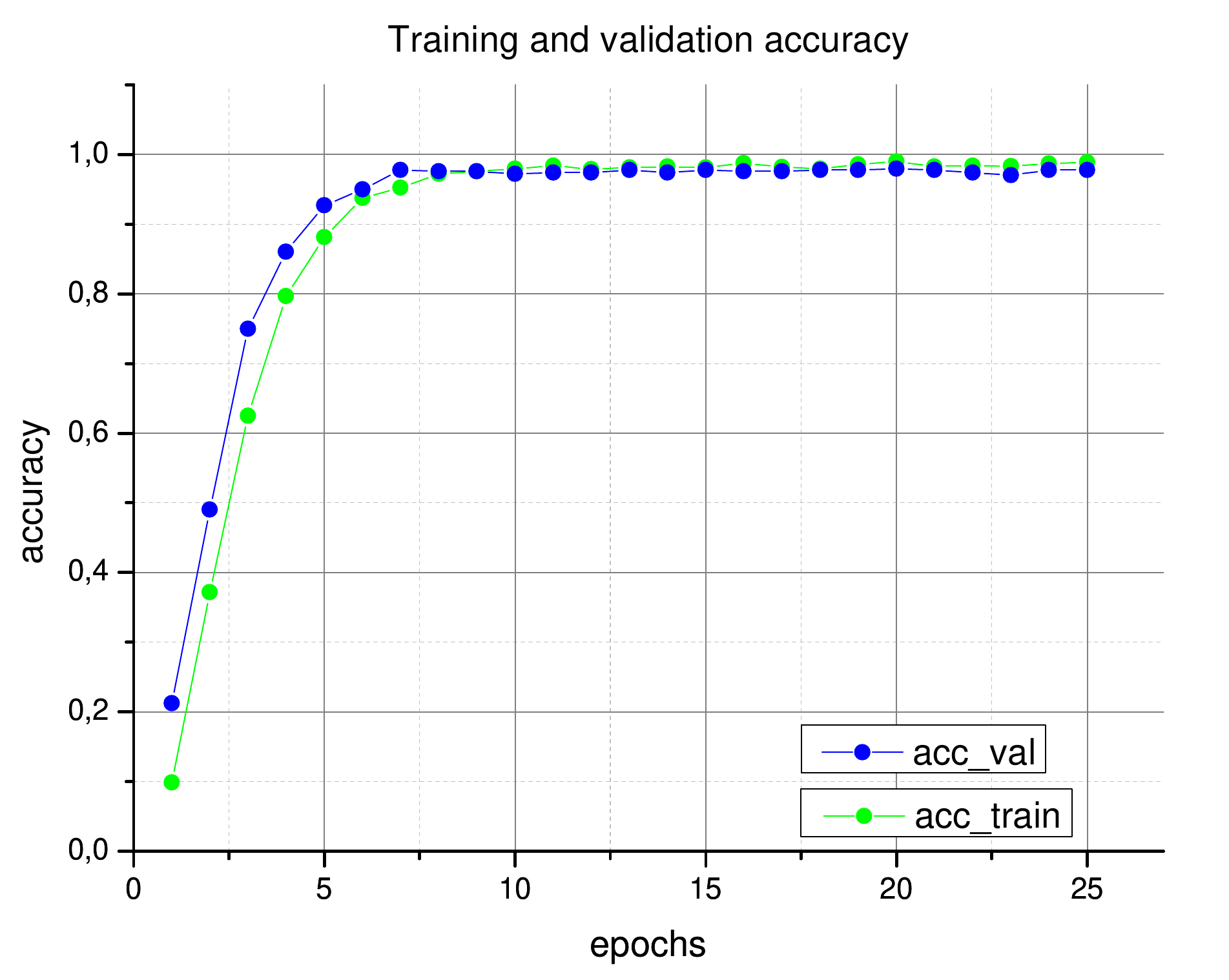}
	\caption{Training and validation accuracy over the first 25 epochs for model 1}
	\label{fig:TrainModel1}
\end{figure}

In our first approach we only use dataset 1 and a PyTorch \cite{PyTorch} implementation of ResNet-18 \cite{Chilamkurthy} pretrained on the ImageNet dataset \cite{ImageNetDataset}. 
For data augmentation we used random horizontal flipping, rotation and cropping images while training. The number of epochs were varied due to hyperparametertuning between \num{10} and \num{60}. The learning rate began at \num{0.001} and decayed every \num{10} epochs by a factor of \num{0.1}. We found that the model converges beginning at epoch \num{25} and no higher validation accuracy can be achieved by longer training time. Figure \ref{fig:TrainModel1} shows the training and validation accuracy per epoch until the model converges.

\subsubsection{Model 2}
Model 2 was first trained on dataset 2 and afterwards on dataset 1.
The model got trained \num{25} epochs on each dataset. The learning rate remained as in model 1.

\subsubsection{Model 3}
To have a basis for evaluating different object detection and classification approaches, we train a SVM classifier on the dataset 2. This classifier is based on the implementation of the scikit-learn library \cite{Scikit}. The pretrained feature extractor net ResNet-101 \cite{ResnetV2.He.2016} was not trained further.

We used a principal component analysis (PCA) to reduce the feature space and the C parameter of the SVM as hyperparameters. The PCA reduces the feature space to \SI{99}{\percent} of the variance of the extracted training features and got applied to the extracted training and validation features. 
We found that the most accurate result was achieved with a C parameter value of \num{50} and no upstream PCA. The training without PCA takes \SI{35}{\second} on a Intel i5 CPU with \SI{8}{GB} RAM. Due to the low test accuracy of model 3 in comparison to model 1 and model 2, we did not apply the final test dataset.

\begin{figure}[h!]
	\centering
	\includegraphics[width=0.5\textwidth]{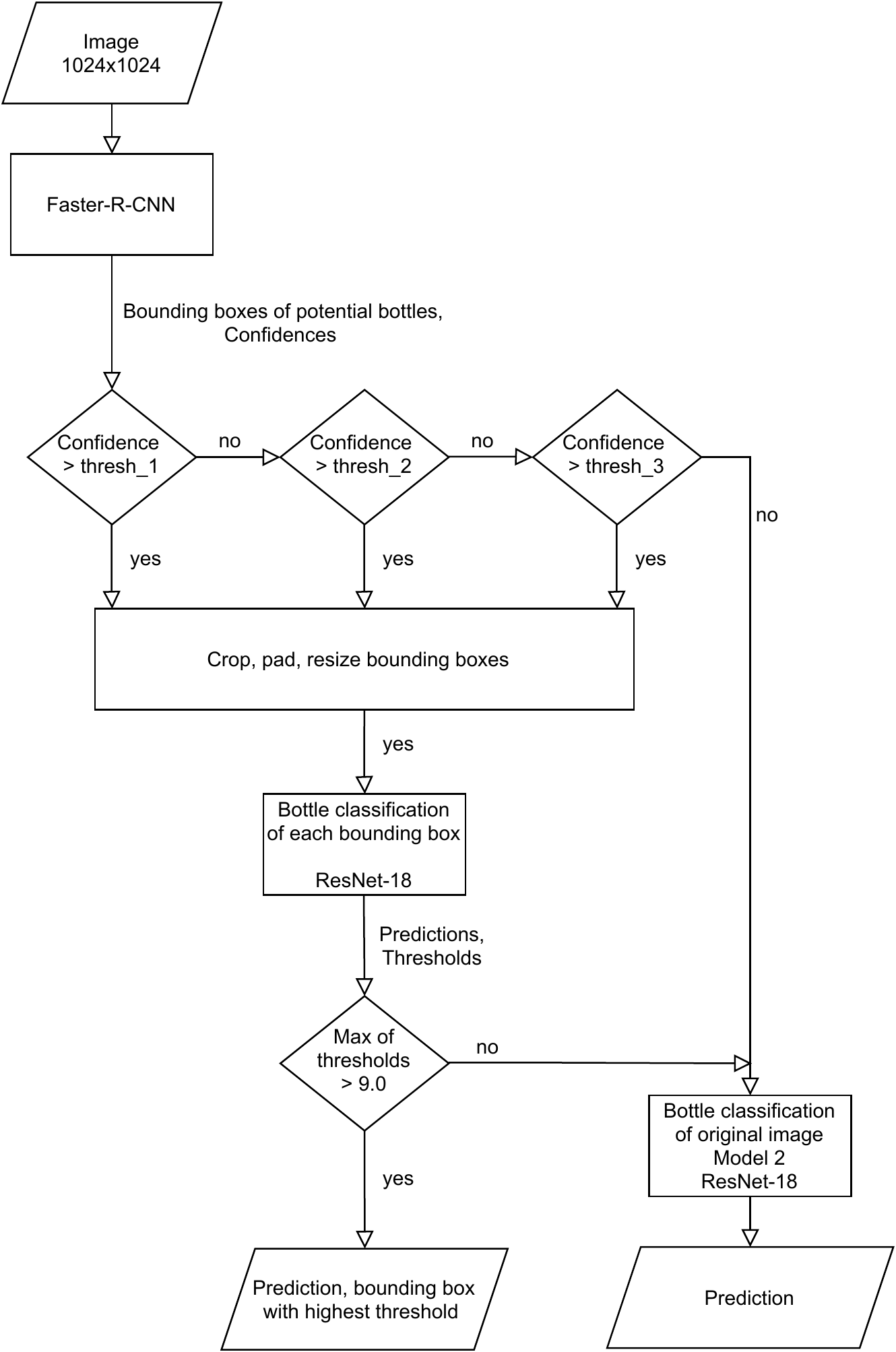}
	\caption{Flowchart of model 4}
	\label{flowchart_model4}
\end{figure}

\subsubsection{Model 4}
This model is based on the object detection and classification approach (see Figure \ref{flowchart_model4}). We used a Faster-R-CNN pretrained on the COCO 2017 dataset with \num{80} classes, including a bottle class. Given the structure of our dataset we know that in each image is exactly one beer bottle to classify. Because of that we tried to receive one bottle from the detector. Using a high threshold of \num{0.7} resulted in missing bottles when it was partly hidden or not the main part of the image. On other images with other types of bottles in it only one false positive bounding box was returned. Because of that we decided to detect all bottles in the image above a certain threshold. We implemented a three step threshold approach to receive at least one beer bottle bounding box from the image and the least number of false positives. We found that the best threshold values for the detector are \num{0.3}, \num{0.1} and \num{0.01}.
For classification of the resulting bounding boxes we used model 1 and trained it on dataset 2. The highest confidence of all detected bounding boxes is taken and compared to a second threshold value of \num{9.0}. Experiments showed that if the confidence is below that threshold the prediction is wrong with a high chance.
If the confidence is below we take the whole image and classify it with our model 2.

\subsubsection{Model 5}
This model has the same architecture as model 4, but uses different model weights and threshold values. The pretrained Faster-R-CNN was finetuned with the BeerBB-1K dataset to detect only beer bottles. With that the number of false positive bounding boxes was reduced and we had to adjust the threshold values to \num{0.5}, \num{0.2} and \num{0.01}. The second threshold value stayed the same with \num{9.0}.

\subsubsection{Model 6} 
This model was created after the DLC ended. It uses the same object detector and classifier as model 4 and model 5 but the layer between them was optimized. We loop over all found bounding boxes starting with the one with the highest confidence. The bounding box is classified and compared to a threshold value of \num{8.0}. If the confidence of the classification is below that threshold we continue with the next bounding box. If no bottle is detected or the confidence of all bounding boxes is below the threshold value the whole image is classified with model 2.

\section{Results and discussion} \label{Results}

The achieved results of all models on dataset 1 are summarized in table \ref{tab:resulttable}.
The simpler approaches, model 1 and model 2 consisting of only a CNN achieved recognition rates of \SI{99.61}{\percent} and \SI{99.52}{\percent}. This was possible due to the pretrained ResNet-18 with \SI{17.6}{M} parameters \cite{b187}. The downside of this model was the need for a GPU or the use of Google Colab to get acceptable training times. Using Google Colab Pro we were able to train model 1 and model 2 within \num{30} minutes.

\begin{figure}[htb!]\includegraphics[width=0.5\textwidth]{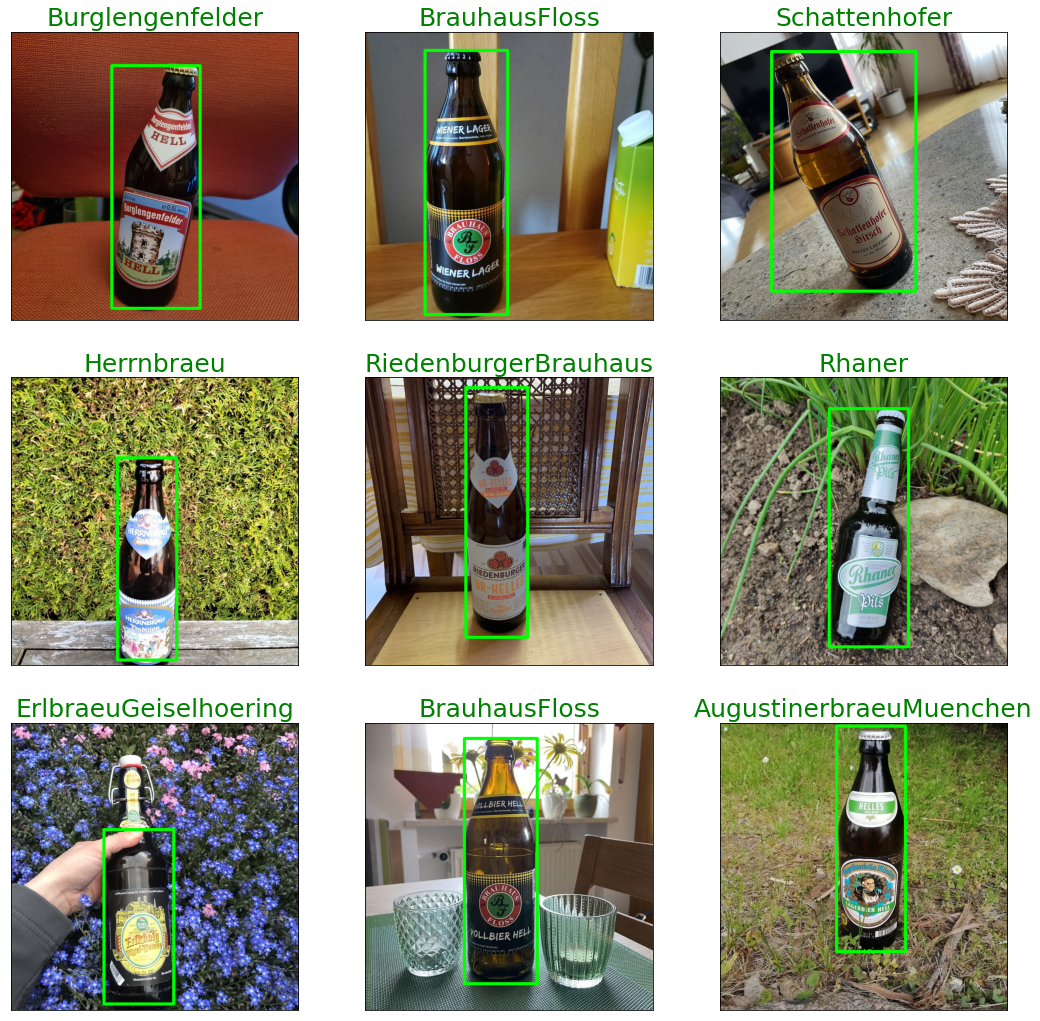} 
\caption{Examples of detected bottles with prediction box and label}
\label{img:detected}
\end{figure}

This was the main reason for trying a different approach solving the DLC. SVMs don't need the same amount of computing power compared to CNNs as described in chapter \ref{Experiments}. However, as seen in table \ref{tab:resulttable}, we could not accomplish the accuracy of our CNN based models on our first approach with SVMs.
Chen et al. \cite{b69} reached a recognition rate of \SI{99.52}{\percent} for an automatic detection of traffic lights using
image processing and SVM techniques. 
So with further investigations it could be possible to reach a higher accuracy than model 3. Because of the limited time of the DLC we focused on the CNNs.
\begin{table}[h]
	\caption{Results of the trained models applied on the training, validation, test and final test dataset splits.}
	\centering
	\begin{tabular*}{\columnwidth}{ @{\extracolsep{\fill}}c c c c c}
		\toprule
		model            &  train acc [\%]   &  val acc [\%]   &   test acc [\%]   & final test acc [\%]   \\ \midrule 
		model 1 & 100 & 99.81 & 99.33 & 99.61\\
		model 2 & 100 & 99.81 & 99.62 & 99.52\\
		model 3 & 99.86 & 61.33 & 55.69 & -\\
		model 4 & 100 & 100 & 100 & 99.87\\
		model 5 & 100 & 100 & 100 & 99.87\\
		model 6 & 100 & 100 & 100 & 100\\
		\bottomrule
	\end{tabular*}
	\label{tab:resulttable}
\end{table}
We got the highest recognition rate for the DLC with model 4 and model 5, who reached both \SI{100}{\percent} accuracy on the test data and \SI{99.87}{\percent} on the final test data.
The identified and assigned bottles including bounding boxes can be seen in figure \ref{img:detected}.

The main idea behind the approach of first object detection and then object classification in the models 4, 5 and 6 is to reduce random noise from the background. The ResNet-18 is only trained on cropped pictures of the bottles. So it focuses on the bottles and minimized the influence of the background.
With this method, we were able to increase the recognition rate.
Model 6 reached \SI{100}{\percent} accuracy on the final test data, but was not taken into account for the DLC.

\section{Conclusion} \label{Conclusion}
In this paper we present six methods for beer bottle classification. Although the dataset is relatively small our models reach a high accuracy through the use of transfer learning. Our final model takes images with a single beer bottle as input and uses a two step approach with object detection and classification with Faster-R-CNN and ResNet-18. It outperforms the accuracy of the classical one step CNN approach and we were able to achieve an accuracy of \SI{100}{\percent} on all provided images. The proposed structure furthermore detects the position of the bottle in the image. In the future our model could be easily converted to detect and classify multiple different beer bottles in an image.

\end{document}